\documentclass{bmvc2k}

\usepackage{float}
\usepackage{verbatim}
\usepackage{enumitem}
\usepackage{booktabs}
\title{Attention Distillation for Learning Video Representations}

\addauthor{Miao Liu}{mliu328@gatech.edu}{1}
\addauthor{Xin Chen}{xinchen@kwai.com}{3}
\addauthor{Yun Zhang}{yzhang467@gatech.edu}{1}
\addauthor{Yin Li}{yin.li@wisc.edu}{2}
\addauthor{James M. Rehg}{rehg@gatech.edu}{1}

\addinstitution{
College of Computing\\
Georgia Institute of Technology\\
Atlanta, United States
}
\addinstitution{
Departments of Biostatistics\\ and of Computer Sciences\\
University of Wisconsin-Madison\\
Madison, United States
}
\addinstitution{
Y-tech \\
Kuaishou Technology\\
Palo Alto, United States
}

\runninghead{Liu, Chen, Zhang, Li, Rehg}{Attention Distillation}

\usepackage{graphicx}
\usepackage{comment}
\usepackage{times}
\usepackage{epsfig}
\usepackage{amsfonts}
\usepackage{multirow}
\usepackage{makecell}
\usepackage{arydshln}
\usepackage{enumitem}
\newcommand{\tablestyle}[2]{\setlength{\tabcolsep}{#1}\renewcommand{\arraystretch}{#2}\centering\footnotesize}
\usepackage{amsmath,amssymb} % define this before the line numbering.
\usepackage{color}

\begin{document}

\maketitle

\begin{abstract}
We address the challenging problem of learning motion representations using deep models for video recognition. To this end, we make use of attention modules that learn to highlight regions in the video and aggregate features for recognition. Specifically, we propose to leverage output attention maps as a vehicle to transfer the learned representation from a flow network to an RGB network. We systematically study the design of attention modules, develop a novel method for attention distillation, and evaluate our method on major action recognition benchmarks. Our results suggest that our method improves the performance of the baseline RGB network by a significant margin while maintains similar efficiency. Moreover, we demonstrate that attention serves a more robust tool for knowledge distillation in video domain. We believe our method provides a step forward towards learning motion-aware representations in deep models and valuable insights for knowledge distillation. Our project page is available at \url{https://aptx4869lm.github.io/AttentionDistillation/}
\end{abstract}
% Moreover, we demonstrate that our attention maps can leverage motion cues in learning to identify the location of actions in video frames.
%-------------------------------------------------------------------------
\section{Introduction}
Action recognition in videos has emerged as a key challenge for deep models. This task requires the understanding of both spatial and temporal cues and the best methods for extracting and fusing them. The two-stream architecture~\cite{simonyan2014two}, exemplified by the I3D model~\cite{carreira2017quo}, has proven to be a effective framework for addressing these challenges. Fusing two modalities of appearance and motion is conceptually appealing, yet it is computationally expensive. A two-stream model can be 100 times slower than its single RGB stream version~\cite{crasto2019mars}. Moreover, learning motion-aware video features from RGB frames remains a challenging problem~\cite{Feichtenhofer_2018_CVPR}. In this context, we address the following research questions: {\it Does a deep model need an explicit flow channel to capture motion patterns? How can we bridge the gap between an RGB stream network and its two stream version without incurring the extra computational cost?} Several previous works have addressed the challenge of learning a video representation that encodes motion information using a single RGB stream~\cite{Tran_2018_CVPR,stroud2020d3d,crasto2019mars,sun2018optical,jiang2019stm}. Our work shares the same motivation, but  pursues a very different approach.

% \begin{figure}[t]
% \centering
% \includegraphics[width=0.68\linewidth]{teaser.pdf} \vspace{-0.6em}
% \caption{RGB and flow networks attend to different aspects of an action, yet both are essential for recognition. Left: Attention maps from RGB and flow streams of I3D~\cite{carreira2017quo} by Grad-Cam~\cite{selvaraju2017grad}. Right: Attention maps from our attention distillation model. Our model jointly infers appearance and motion attention from only RGB frames, improves the performance of the RGB stream, and is significantly more efficient than two-stream models.} \vspace{-1.5em}
% \label{fig:teaser}
% \end{figure}

We present a novel video representation learning method called \emph{attention distillation}. Our method makes use of an explicit probabilistic attention model, and leverages motion information available at training time to predict the motion-sensitive attention features from a single RGB stream. In addition to their utility in visualizing and understanding learned feature representations, we argue that attention models provide an attractive vehicle for mapping between sensing modalities in a task-sensitive way. Once learned, our model requires only RGB frames as inference inputs, and jointly predicts appearance and motion attention maps for action recognition. We conduct extensive experiments and demonstrate that our attention distillation enables more accurate action recognition across several video datasets, while remaining very efficient.

Our main contributions are summarized as follows: 
\begin{itemize}
    \item  We propose a novel method for learning motion-aware video representations from RGB frames. Our method distills motion knowledge into an RGB network by mimicking the attention map of a reference flow network. \vspace{-0.6em}
    \item Our method achieves consistent improvements of $\sim$ 1\% across major datasets, including UCF101~\cite{Soomro2012UCF101AD}, HMDB51~\cite{kuehne2011hmdb}, EGTEA~\cite{Li_2018_ECCV}, and 20BN-V2~\cite{mahdisoltani2018fine}, with almost no extra computational cost. \vspace{-0.6em}
    \item We study different choices of attention modules for action recognition, and demonstrate that attention serves as a more robust vehicle for knowledge distillation in comparison to previous feature distillation methods.\vspace{-0.6em}
\end{itemize}

\section{Related Works}
\subsection{Action Recognition} 
Action recognition is well studied in computer vision~\cite{poppe2010survey}. Recent efforts focus on developing novel deep models for action recognition. For example, recent works~\cite{tran2015learning,Hara_2018_CVPR,carreira2017quo} proposed to make use of 3D convolutional networks to capture spatio-temporal features beyond a single frame. However, their performance using video frames alone falls far behind their two stream versions~\cite{carreira2017quo}. Our work seeks to address the problem of recovering the motion cues encoded in videos from RGB frames alone. There are several recent attempts in this direction. Bilen et al.\ \cite{bilen2018action} proposed a dynamic image network that makes use of the parameters of a ranking machine that captures the temporal evolution of the video frames. Ng et al.\ \cite{ng2016actionflownet} proposed to jointly predict action labels and flow maps from video frames using multi-task learning. This idea is extended by Fan et al.\ \cite{fan2018end}, where they fold the TV-L1 flow estimation~\cite{perez2013tv} into their TVNet. Without using flow, Tran et al.\ \cite{Tran_2018_CVPR} demonstrated that factorized 3D convolutions (2D spatial convolution and 1D temporal convolution) can facilitate the learning of spatio-temporal features. A similar finding was also presented by Xie et al.\ \cite{Xie_2018_ECCV}. Our method shares the same motivation as these approaches, yet takes a vastly different route. We explore attention mechanisms for video recognition, and propose to distill the predicted attention from a flow network to an RGB network.

\subsection{Knowledge Distillation}
Our attention distillation method is inspired by knowledge distillation, first proposed by~\cite{bucilu2006model} for model compression and further popularized by~\cite{hinton2015distilling}. Some recent works~\cite{gupta2016cross,garcia2018modality,luo2018graph} explored knowledge distillation across modalities. The most relevant works are~\cite{stroud2020d3d,crasto2019mars}. They both addressed the challenge of video representation learning via knowledge distillation. They assume that the reference flow network has better performance than the RGB stream network, and seek to regularize the learning of the RGB stream by distilling features from the flow network to the RGB stream. However, for recently developed large scale video datasets (e.g., Kinetics~\cite{kay2017kinetics}, Charades~\cite{sigurdsson2016hollywood}) and egocentric video datasets (e.g., EGTEA~\cite{Li_2018_ECCV} and EPIC-Kitchens~\cite{damen2018scaling}), the RGB stream has better performance than the flow stream. Feature distillation methods also suffer from the potential threat of ``overwriting'' the features from RGB stream with the features from flow stream. This is related to the pitfall of catastrophic forgetting~\cite{french1999catastrophic}.  In contrast, we propose to distill attention maps-- which are indicators of important regions for recognition. This design choice stems from the key challenge of video representation learning--motion is substantially different from appearance and both modalities are important for recognition. Our experimental results in Sec.~\ref{sec:exp} demonstrate that our method can overcome the disadvantages of previous feature distillation methods.

\subsection{Attention for Recognition} 
Attention has been widely used for visual recognition. We focus on selective visual attention that highlights discriminative regions. This is very different from the recent efforts on self-attention, i.e., self-similarity~\cite{vaswani2017attention,Girdhar_2019_CVPR,zhang2018self}. Recently, selective attention has been explored in deep models for object recognition~\cite{mnih2014recurrent} and image captioning~\cite{xu2015show}. Attention enables these models to ``fixate'' on image regions, where the decision is made based on a sequence of fixations. Several attention mechanisms are proposed for deep models. For example,~\cite{sharma2015action} integrated soft attention in LSTMs for action recognition.~\cite{li2018videolstm} further extends~\cite{xu2015show} into videos. Specifically, they combined LSTMs with motion-based attention to infer the location of the actions.~\cite{Girdhar_17b_AttentionalPoolingAction} modeled top-down and bottom-up attention using bilinear pooling.~\cite{wang2017residual} proposed a residual architecture for soft attentions. Our previous work~\cite{Li_2018_ECCV, li2020eye} considered attention as a probabilistic distribution for egocentric action recognition. Our recent work~\cite{liu2019forecasting} made use of motor attention for action anticipation. In this paper, we demonstrate that a useful probabilistic attention model can be obtained without access to a prior distribution from human gaze data. We also provide a systematical study of the utility of probabilistic attention model in action recognition. 

\section{Distilling Motion Attention for Actions}

In this section, we present our method of attention distillation. We start with an overview of the key ideas, followed by a detailed description of the components in our method. Finally, we describe our network architecture and discuss the implementation details.

\subsection{Overview}
For simplicity, we consider an input video with a fixed length of $T$ frames. Our method can easily generalize to multiple videos, e.g., for mini-batch training. We denote the input video as $x=\{x^1, x^2, ..., x^T\}$, where $x^t$ is a frame of resolution $H\times W$ with $t$ as the frame number. Given $x$, our goal is to predict a video-level action label $y$. We leverage the intermediate output of a 3D convolutional network $\phi$ to represent $x$. This is given by a 4D tensor $\phi(x)$ of the size $T_\phi \times H_\phi \times W_\phi \times C_\phi$. $C_\phi$ is the feature dimension of 3D grids $T_\phi \times H_\phi \times W_\phi$ from the video $x$. Our method consists of three key components:

\noindent \textbullet\ \textbf{Attention Generation}. Our model first predicts an attention map $\mathcal{A}$ based on $\phi(x)$ using the attention mapping function $F_{\mathcal{A}}$. $\mathcal{A}$ is a 3D tensor of size $T_\phi \times H_\phi \times W_\phi$. Moreover, $\mathcal{A}$ is normalized within each temporal slice, i.e., $\sum_{w,h} \mathcal{A}(t,w,h)=1$. $\mathcal{A}$ is thus a sequence of 2D attention maps $\mathcal{A}(t)$ defined over $T_\phi$ steps. 

\noindent \textbullet\ \textbf{Attention Guided Recognition}. Based on the attention map $\mathcal{A}$ and the feature map $\phi(x)$, our model further applies a recognition module $F_{\mathcal{R}}$ to predict the action label $y$. Specifically, this module uses $\mathcal{A}$ to selectively pool features from $\phi(x)$, followed by a classifier that maps the result feature vectors to the action label $y$.

\noindent \textbullet\ \textbf{Attention Distillation}. To regularize the learning, we assume that $\mathcal{A}$ will receive supervision from a teacher model that outputs a reference attention map $\mathcal{\tilde{A}}$. The teacher model comes from the flow stream and is equipped with the same attention module for recognition.

\begin{figure*}[t]
\centering
\includegraphics[width=0.8\linewidth]{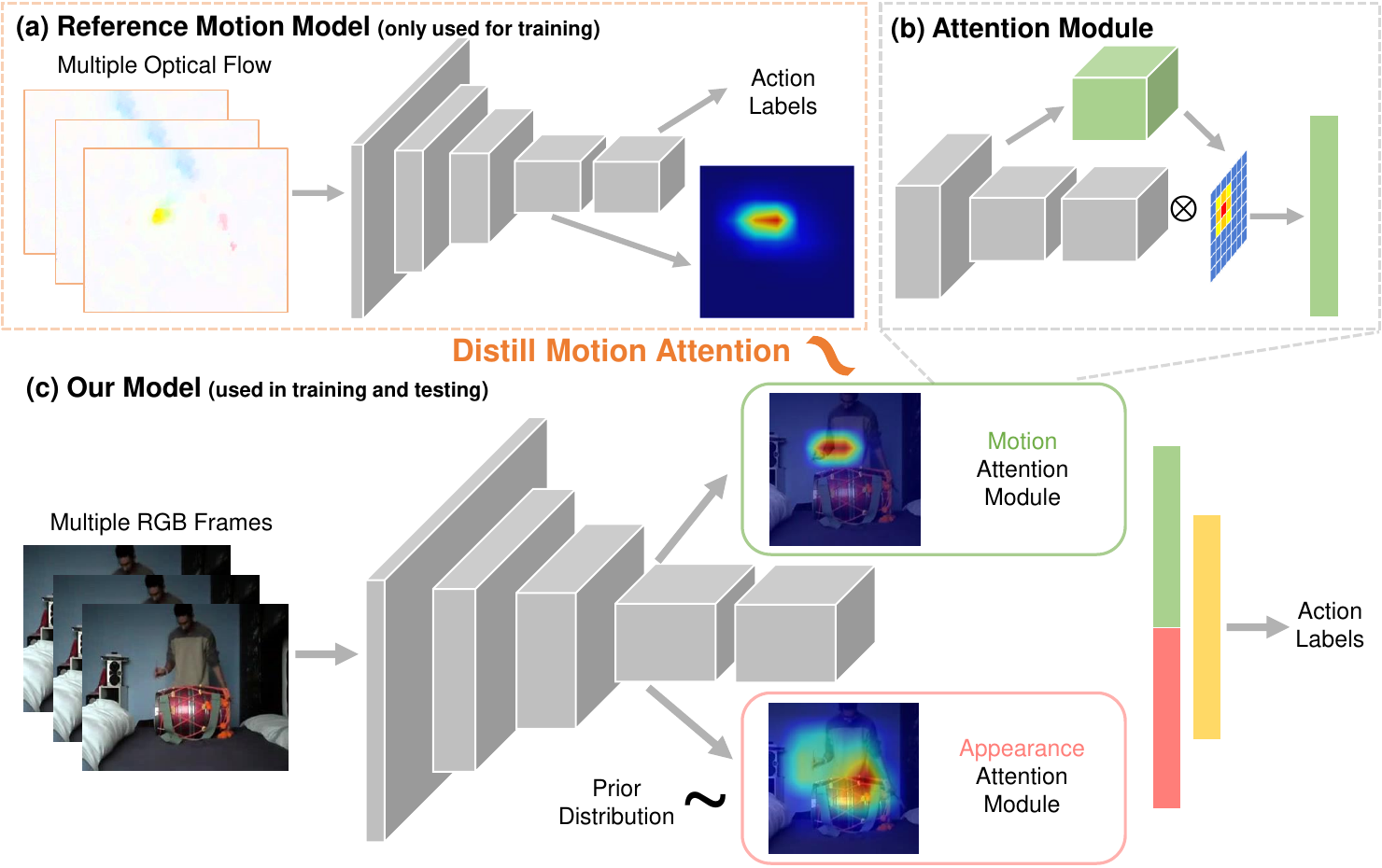}
\caption{Overview of our method. Our model (c) takes multiple RGB frames as inputs and adopts a 3D convolutional network as the backbone. It outputs two attention maps using the attention module (b), based on which the action labels are predicted. The motion map is learned by mimicking the attention from a reference flow network (a). The appearance map is learned to highlight discriminative regions for recognition. These two maps are used to create spatio-temporal feature representations from video frames for action recognition.}
\label{fig:flowchart}
\end{figure*}
Fig.\ \ref{fig:flowchart} presents an overview of our method. Our model takes multiple video frames $x$ as inputs, and learns to predict two attention maps based on $\phi(x)$: $\mathcal{A}^M$ for motion attention and $\mathcal{A}^A$ for appearance attention. Based on these two maps, the model further aggregates visual features that will be passed into the final recognition sub-network. During training, we match $\mathcal{A}^M$ to the attention map $\mathcal{\Tilde{A}}^M$ from the reference flow network. For testing, only the input video is required for recognition. Our model also outputs two attention maps that can be used to visualize and diagnose recognition performance. We now detail the design of our key components.

\subsection{Attention Generation}
We explore two different approaches for generating an attention map from the features $\phi(x)$, including soft attention~\cite{wang2017residual} and its probabilistic version~\cite{Li_2018_ECCV}.

\noindent \textbf{Soft Attention}. Attention maps can be created by a linear function of $w_a \in R^{C_\phi}$ over the feature map $\phi(x)$,
\begin{equation}
\small
    F_{\mathcal{A}}(\phi(x)) = softmax(w_a * \phi(x)),
\end{equation}
where $*$ is the 1x1 convolution on 3D feature grids. Softmax is applied on every time slice to normalize each 2D map.

\noindent \textbf{Probabilistic Soft Attention}. An alternative approach is to further model the distribution of linear mapping outputs as discussed in~\cite{Li_2018_ECCV}, namely 
\begin{equation}
\small
    \mathcal{A} \sim p(\mathcal{A}) = softmax(w_a * \phi(x))
\end{equation}
where we model the distribution of $A$. During training, an attention map can be sampled from $p(\mathcal{A})$ using Gumbel Softmax trick~\cite{jang2016categorical,maddison2016concrete}. We follow~\cite{Li_2018_ECCV} to regularize the learning by adding additional loss term of 
\begin{equation} \label{eq:reg_atten}
\small
    \mathcal{L}^R = \sum_t KL\left[\mathcal{A}(t) || U\right], 
\end{equation}
where $KL[\cdot]$ is the Kullback-Leibler divergence and $U$ is the 2D uniform distribution ($H_\phi \times W_\phi$). This term matches each time slice of the attention map to the prior distribution. It is derived from variational learning and accounts for (1) the prior of attention maps and (2) additional regularization by spatial dropout~\cite{Li_2018_ECCV}. During testing, we directly plug in $p(\mathcal{A})$ (the expected value of $\mathcal{A}$) for approximate inference.

Note that for both approaches, we restrict $F_{\mathcal{A}}$ to a linear mapping without a bias term. In practice, this linear mapping avoids the trivial solution of generating a uniform attention map by setting $w$ to all zeros. This all-zero solution almost never arises during training when using a proper initialization of $w$.

\subsection{Attention Guided Recognition}
Our recognition module makes use of an attention map $\mathcal{A}$ to select features from $\phi(x)$. Again, we consider two different models for the attention guided recognition. 

\noindent \textbf{Attention Pooling}. Inspired by~\cite{wang2017residual,liu2019end}, we design the function $F_{\mathcal{R}}$ as 
\begin{equation}
\label{eq:direct_atten}
\small
    \Tilde{y} = F_{\mathcal{R}}(\phi(x), \mathcal{A}) = softmax\left( W_r^T (\mathcal{A} \otimes \phi(x)) \right)
\end{equation}
where $\otimes$ denotes the tilted multiplication $\mathcal{A} \otimes \phi(x) = \sum_{t,h,w} \mathcal{A}(t, h, w) \phi(x)_{t,h,w,c}$. This operation is equivalent to weighted average pooling with the weights shared across all channels.
% followed by linear classifiers $W_r$ with softmax normalization. 

\noindent \textbf{Residual Connection}. Using the attention map to re-weight features helps to filter out background noise, yet may also increase the potential risk of missing important foreground features. This drawback was discussed in~\cite{wang2017residual}. We follow their solution of using a residual connection to the attention map, given by 
\begin{equation}
\label{eq:res_atten}
\small
    \Tilde{y} = F_{\mathcal{R}}(\phi(x), \mathcal{A}) = softmax\left( W_r^T ( (\mathcal{A}+I) \otimes \phi(x) ) \right),
\end{equation}
where $I$ is a 3D tensor of all ones. Intuitively, this operation further adds average pooled features to the representation before the linear classifier. By adding the residual term, the features learned by the network are preserved. 

\subsection{Attention Distillation}
The key to our approach lies in the use of attention distillation during training. Specifically, we assume that a reference flow network is given as the teacher network. The teacher model also uses an attention mechanism for recognition. Moreover, its motion attention map $\mathcal{\Tilde{A}}^M$ is used as additional supervisory signal for training our RGB network. This RGB network is thus the student model that mimics the motion attention map. With probabilistic attention modeling, the imitation of the attention maps is enforced by using the loss
\begin{equation}
\label{eq:distill}
\small 
    \mathcal{L}^{\mathcal{A}} = \sum_t\  KL\left[ \mathcal{A}^M(t)||\mathcal{\Tilde{A}}^M(t) \right].
\end{equation}
This loss minimizes the distance between the attention maps at every time step $t$. In our implementation, our teacher flow network is trained with the same attention mechanism. Once trained, the weights of the teacher model remain fixed during the learning of the student model. At testing time, only the student model (RGB network) is used for inference. 

\subsection{Our Full Model}
Putting everything together, we summarize our full model with probabilistic soft attention and attention distillation. Specifically, our model estimates the two probabilistic attention maps $\mathcal{A}^M \sim F_{\mathcal{A}}^M(\phi(x))$ (motion) and $\mathcal{A}^A \sim F_{\mathcal{A}}^A(\phi(x))$ (appearance). These maps are further used to predict the action labels. This is given by 
\begin{equation}
\small
    \Tilde{y} = F_{\mathcal{R}}^M(\phi(x), \mathcal{A}^M) + F_{\mathcal{R}}^A(\phi(x), \mathcal{A}^A)
\end{equation}
where each $F_{\mathcal{R}}$ follows Eq~\ref{eq:direct_atten}. We use equal weighting for $F_{\mathcal{R}}^M$ and $F_{\mathcal{R}}^A$. We found that tuning the weights has negligible effect on the performance in practice.

\noindent \textbf{Loss Function}. Our training loss is defined as
\begin{equation}
\small
\begin{split}
    \mathcal{L} = CE(\Tilde{y}, y) &+ \lambda_1 \sum_t\  KL\left[ \mathcal{A}^M(t)||\mathcal{\Tilde{A}}^M(t) \right] + \lambda_2 \sum_t KL\left[ \mathcal{A}^A(t)||U \right],
\end{split}
\end{equation}
where $CE$ is the cross entropy loss between the predicted labels $\Tilde{y}$ and the ground-truth $y$. Thus, the loss consists of three terms. The first cross entropy term is to minimize the error for classification. The second KL term (from Eq.\ \ref{eq:distill}) enforces that the motion attention $\mathcal{A}^M$ should mimic the attention map $\mathcal{\Tilde{A}}^M$ from the reference flow network. Finally, the third KL term (from Eq.\ \ref{eq:reg_atten}) regularizes the learning of the appearance attention. The coefficients $\lambda_1$ and $\lambda_2$ are used to balance the three terms. We choose $\lambda_1=1$  and $\lambda_{2} = 1/(T_\phi\times W_\phi\times H_\phi)$.

\subsection{Implementation Details}
\noindent \textbf{Network Architecture}.\ 
Our model uses I3D network~\cite{carreira2017quo} as the backbone. I3D has five 3D convolution blocks, and three of them are composed of multiple Inception Modules. For all attention modules, the intermediate feature $\phi$ is obtained from the outputs of the 4th convolutional block. The attention map is used to select the final network feature from the last Inception module of the 5th convolutional block.

\noindent \textbf{Data Preparation}.\
We down-sample all frames to $320\times256$ with a frame rate of 24 Hz. For training, we compute optical flow using TV-L1~\cite{perez2013tv}. We apply several data augmentation techniques, including random flipping, cropping and color perturbation to prevent over-fitting. Our model takes 24 consecutive frames as inputs, and all input frames are cropped to $224\times224$ for training. For testing, we evaluate our model on full resolution clips ($320\times256$) and aggregate scores from all clips to produce the video-level results.

\noindent \textbf{Training and Inference Details}.\
All of our models are trained using SGD with momentum of 0.9. The weights are initialized from Kinetics pre-trained models provided by the authors of~\cite{carreira2017quo}. Our models are trained with a batch size of 64 on 4 GPUs. The initial learning rate is 0.01 with a decay rate of 10 when the loss starts to saturate. We set weight decay to 4e-5 and enable batch norm~\cite{ioffe2009batch}. We also adopt dropout rate 0.7. {\it At inference time our model does not need optical flow}, and runs at the same speed as the RGB network. 

\section{Experiments}
\label{sec:exp}
We now present our experiments and results. We start with a systematical evaluation of attention guided action recognition, followed by our main results on several public datasets. 

\subsection{Datasets and Metrics}
We make use of four action recognition datasets for our experiments: UCF101, HMDB51, EGTEA Gaze+ (egocentric videos) and 20BN-V2. UCF101~\cite{Soomro2012UCF101AD} has 13,320 videos from 101 action categories. HMDB51~\cite{kuehne2011hmdb} includes 6,766 videos from 51 action categories. EGTEA~\cite{Li_2018_ECCV,li2020eye} contains 10,321 videos from 106 action categories. We evaluate mean class accuracy and report the results using the first split of these three datasets. 20BN-V2~\cite{mahdisoltani2018fine} has over 220K videos from 174 fine-grained action categories. We use their training and validation split, and report top-1/top-5 accuracy.

\subsection{Attention Guided Action Recognition}
We start from an ablation study of attention-guided action recognition. Specifically, we evaluate different combinations of attention modules and compare their results to those from models without attention. Our experiments show that the proper design of the attention mechanism can consistently improve the performance of action recognition across multiple datasets. We now present our baselines and results. 

\noindent \textbf{Baselines}. We consider the different combinations of how the model generates attention maps (Soft vs.\ Probabilistic Attention) and how the attention maps are used for recognition (Attention Pooling vs.\ Residual Connection). In addition, we also show how the approach to combining motion attention and appearance attention affects the recognition performance. The valid combinations include the following:\vspace{-0.5em}
\begin{itemize}[leftmargin=*]
    \item \textbf{Soft-Atten} combines soft attention and attention pooling for recognition similar to~\cite{liu2019end}. \vspace{-0.5em}
    \item \textbf{Soft-Res} is the residual attention in~\cite{wang2017residual} that adds residual connection to Soft-Atten.\vspace{-0.5em}
    \item \textbf{Prob-Atten} combines probabilistic attention with attention pooling as in~\cite{Li_2018_ECCV}.\vspace{-0.5em}
\end{itemize}
We note that the combination of Prob+Res is invalid, as it violates the probabilistic modeling of attention. In practice, we also found its training to be unstable. Therefore, we report the results of three valid designs for both RGB and flow stream and the vanilla I3D models (our backbone) using the same input sequence length (24 frames) in Table~\ref{table:ablation}. Adding attention to the backbone recognition network almost always improves the performance. Importantly, Soft-Res decreases the performance of RGB stream on HMDB51 and Soft-Atten decreases the performance of RGB stream on HMDB51 and UCF101. More interestingly, Prob-Atten is the most robust design choice, despite the lack of human gaze as a supervisory signal as in~\cite{Li_2018_ECCV}. Across all of the modalities and datasets, Prob-Atten can consistently improve the recognition accuracy (+$0.3\%/0.4\%/1.8\%$) for the RGB stream and (+$0.9\%/0.5\%/2.1\%$) for the flow stream. The performance boost from the attention module is larger for the flow stream in comparison to the RGB stream. Moreover, attention modules provide more significant boost for egocentric actions (EGTEA). We conjecture that the explicit modeling of attention helps to suppress background objects in first person video. 
\begin{table}[t]
\footnotesize 
% \footnotesize
\centering
\tablestyle{2pt}{1.0}
\setlength{\tabcolsep}{5pt} % Default value: 6pt
\renewcommand{\arraystretch}{1} % Default value: 1
\begin{tabular}{c|ccc}
Method                                          & UCF101        & HMDB51       & EGTEA \\ \hline 
\makecell{Flow I3D}  & 94.0  & 73.9  & 38.3 \\
\makecell{Flow Soft-Atten}      & 94.7  & 74.1  & 39.1 \\
\makecell{Flow Soft-Res}        &\textbf{95.2} &\textbf{74.4} & 39.5  \\													               
\makecell{Flow Prob-Atten}      & 94.9  &74.2   &\textbf{40.4}  \\ 
\hline
\makecell{RGB I3D}  & 94.8   & 70.9  & 47.3   \\
\makecell{RGB Soft-Atten}      & 94.7   & 70.8  & 48.6   \\
\makecell{RGB Soft-Res}        & 94.9   & 70.1  & 48.6  \\                    
\makecell{RGB Prob-Atten}  & \textbf{95.1} & \textbf{71.3}  & \textbf{49.1} \\
\end{tabular}\vspace{0.1em}
\caption{Evaluations of attention modules. We compared 3 different design choices with RGB/flow stream on three datasets. Prob-Atten provides a consistent performance boost on both streams and across datasets.}
\label{table:ablation}
\end{table}

\subsection{Attention Distillation for Action Recognition}

We now evaluate our method of attention distillation. In this setting, we assume a reference flow network with attention module is given at training time. We attach motion and appearance attention modules to our RGB backbone. Both attention heads follow the same attention module design as the reference network. The flow attention is asked to mimic the motion attention map from the reference flow network. {\it During testing our model does not need optical flow}, and runs at the same speed as the RGB network (about 100 times faster than a two stream network~\cite{crasto2019mars}). We present our results for action recognition, and contrast our method with feature distillation methods~\cite{Zagoruyko2017AT,crasto2019mars}.

\begin{table}[t]
\footnotesize
\centering
\tablestyle{2pt}{1.0}
\setlength{\tabcolsep}{6pt} % Default value: 6pt
\renewcommand{\arraystretch}{1.05} % Default value: 1
\begin{tabular}{c|cc}
Method                                          & UCF101        & HMDB51        \\ \hline                                                                      
\makecell{Dynamic Image~\cite{bilen2018action} }                               & 90.6             & 61.3              \\
\makecell{ActionFlowNet~\cite{ng2016actionflownet}}                   & 83.9              & 56.4            \\	
\makecell{TVNet~\cite{fan2018end} }                                   & 94.5              & 71.0              \\  

\makecell{I3D RGB\text{*}~\cite{carreira2017quo} }         & 94.8              & 70.9            \\
\makecell{FeatMatch~\cite{Zagoruyko2017AT} }         & 94.3              & 70.7            \\
\makecell{MARS~\cite{crasto2019mars}}                                   & 94.6              & \textbf{72.3}              \\  
\makecell{Ours (Prob-Distill)}                                                 & \textbf{95.7}     & \textbf{72.0}            \\
\hdashline
\makecell{Two Stream ResNeXt~\cite{crasto2019mars}}  & 95.6   & 74.0  \\
\makecell{MARS+Flow ResNeXt~\cite{crasto2019mars}}  & 94.9   & 74.5  \\
\makecell{Two Stream I3D\text{*}}  & 96.7   & 74.8  \\
\makecell{Prob-Distill+Flow I3D\text{*}}      & \textbf{97.4}   & \textbf{75.7}  
\end{tabular}\vspace{0.1em}
\caption{Action recognition results on UCF101 and HMDB51 datasets. We compare the results of our model with previous works. Our model outperforms state-of-the-art methods that use only RGB stream and the same input sequence length by $\sim1\%$. \text{*}For fair comparison, we report results of I3D models that use 24 frames as inputs--the same as our model.}
\label{table:action}
\end{table}

\begin{table}[h]
\centering
\tablestyle{2pt}{1.0}
\setlength{\tabcolsep}{6pt} % Default value: 6pt
\renewcommand{\arraystretch}{1} % Default value: 1
\footnotesize
\begin{tabular}{c|cc}
Method                                 & Top-1/Top-5 Acc  & Temporal Footprints  \\ \hline 
TRN RGB~\cite{zhou2017temporal}        & 48.8 / 77.6      & 5 sec        \\
TRN RGB+Flow~\cite{zhou2017temporal}   & \textbf{55.5} / \textbf{83.1}      & 5 sec        \\ \hline
I3D RGB               & 47.3 / 76.1      & 1 sec     \\
I3D Flow                   & 46.7 / 75.9      & 1 sec      \\ 
Ours (Prob-Distill)                             & \textbf{49.9} / \textbf{79.1}      & 1 sec        \\
\hdashline
Two Stream I3D                  & 53.7 / 82.5      & 1 sec      \\ 
Prob-Distill+Flow I3D                            & \textbf{54.6} / \textbf{83.0}      & 1 sec        \\
\end{tabular}\vspace{0.1em}
\caption{Action recognition results on on 20BN-V2 dataset~\cite{mahdisoltani2018fine}. Our model achieves the best performance among networks that uses RGB frames. Fusing our model with a flow network also outperforms two stream baseline by a significant margin.\vspace{-1em}}
\label{table:20bn}
\end{table}

\noindent \textbf{Impact of Attention Distillation}. Table~\ref{table:action} compares our results with previous methods on UCF101/HMDB51. We denote our models using Prob-Atten for distillation as {\it Prob-Distill}. Prob-Distill outperforms all previous state-of-the-art methods of motion representation learning. Specifically, our results are at least $1.2\%$ better than previous state-of-the-art methods for learning motion-aware video representations from RGB frames, including Dynamic Image~\cite{bilen2018action}, ActionFlowNet~\cite{ng2016actionflownet} and TVNet~\cite{fan2018end}. Our model also outperforms MARS~\cite{crasto2019mars}, our direct competitor, by $0.9\%$ on UCF101 and performs on-par with MARS on HMDB51 when using a similar sequence length, despite the fact that MARS uses a stronger backbone network. It is worth noting that this performance boost is significant for action recognition. In contrast, with 50 more layers, ResNet101 is only $0.7\%$ better than ResNet50 on HMDB51~\cite{hara2018can}. Moreover, Prob-Distill also outperforms another feature distillation method -- FeatMatch~\cite{Zagoruyko2017AT} by a significant margin ($+1.4\%/1.3\%$ on UCF101/HMDB51). These results support our argument that \emph{distilling attention maps is more robust than distilling network features for motion representation learning}. Finally, a late fusion of our model with a reference flow network helps to further boost the performance. 

Table~\ref{table:20bn} presents our results on a large scale dataset---20BN-V2. With 1/5 of the temporal receptive field as TRN~\cite{zhou2017temporal}, our model with RGB frames outperforms TRN RGB by $1.1\%/1.5\%$ in top-1/top-5 accuracy. And our method improves the backbone by $2.6\%/3.0\%$ in top-1/top-5 accuracy. Further fusion of our model with a flow network improves the results by a large margin (+4.7\%), again outperforming the two stream baseline.  

\noindent \textbf{Learning from a Weak Flow Network}. Crasto et al.\ \cite{crasto2019mars} pointed out that their model ran into a failure mode when the reference flow network has worse performance than the RGB network. To support our claim that attention distillation can leverage a flow-based teacher network even when the flow network does not provide strong baseline performance, we report the results of our model on the EGTEA Gaze+ dataset. Due to severe ego-motion, flow-based models are less effective than RGB models on this dataset. For instance, I3D Flow is $9\%$ worse than I3D RGB ($38.3\%$ vs.\ $47.3\%$). Despite a much weaker teacher model, Prob-Distill achieves $49.5\%$, outperforming the best attention-based I3D models for both RGB (Prob-Atten $49.1\%$) and Flow (Prob-Atten $40.4\%$). This indicates that even with a weak teacher model, our proposed method is a robust approach to video representation learning.

\noindent \textbf{Distillation without Forgetting}. Feature distillation might ``overwrite'' the features from RGB stream with the features from flow stream. This is evidenced by the result that fusing MARS with reference flow stream network lags behind the two stream version of the network (MARS + Flow ResNeXt vs. Two Stream ResNeXt in Table~\ref{table:action}). In contrast, fusing our Prob-distill model with a reference flow model (Prob-Distill + Flow I3D in Table~\ref{table:action}) further improves the accuracy and outperforms the two-stream I3D model (+$0.5\%$ on UCF101, +$0.7\%$ on HMDB51 and +$0.9\%$ on 20BN-V2). These results indicate that our attention distillation model does not simply copy the feature from the reference flow network, as the distilled RGB model can still preserve meaningful appearance features. 

% \noindent \textbf{Fusion Optimization}. Previous work~\cite{Feichtenhofer_2018_CVPR} pointed out that the late fusion process of two-stream network is far from optimal. In Table~\ref{table:action} and~\ref{table:20bn}, we show that fusing Prob-Distill with flow stream further boosts the performance of two-stream network by a notable amount (+$0.7\% / 0.9\% / 0.9\%$ on UCF101 / HMDB51 / 20BN-V2). Part of the performance gain comes from the attention module. Yet, these results also suggest that our proposed attention distillation method can regularize the learning of video representation, thereby preventing meaningful information being suppressed during fusion.

Note that our single-stream (Prob-Distill) results still lag behind the two stream networks when using the same input sequence length (Two Stream I3D*). This gap reveals that our model does not fully capture the concepts of motion that are encoded in the two stream networks. Nonetheless, we believe that our model provides a key step forward for learning motion-aware representations from RGB frames. Note that some most recent works achieved better performance on the benchmark datasets using more advanced network  structure~\cite{hara2018can,Wang_2018_CVPR,Xie_2018_ECCV,Piergiovanni_2019_CVPR}, additional features~\cite{choutas2018potion}, or a longer temporal footprint~\cite{carreira2017quo,crasto2019mars}. In this context, our work provides a novel method for learning video representations and a robust strategy for knowledge distillation. In supplementary materials, we provide additional analysis to show how the learned attention maps help to localize the spatial extent of actions. We also demonstrate how motion information is encoded in the distilled model.

\begin{figure*}[t]
\centering
\includegraphics[width=0.88\linewidth]{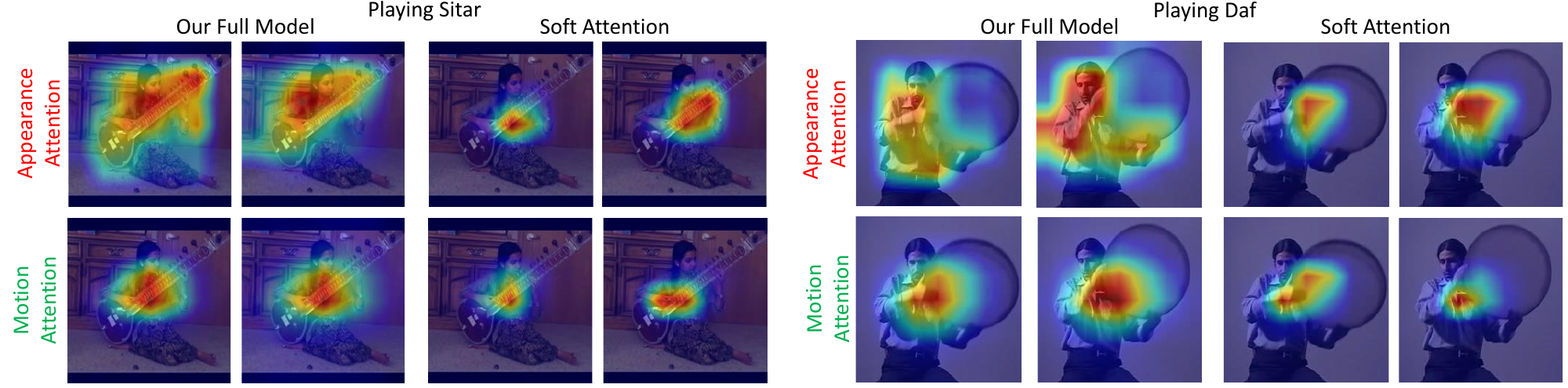}\vspace{-0.6em}
\caption{Visualization of attention maps (Ours vs.\ Soft-Atten using the same I3D backbone). For each video clip, we re-interpolate the attention maps and plot them on the first and last frame. Red regions indicate higher value of attention. Our model produces appearance and motion attention maps that are qualitatively different and index key action regions.}\vspace{-1.5em}
\label{fig:vis}
\end{figure*}

\noindent \textbf{Visualization of Attention Maps}. To better understand our model, we visualize both motion and appearance attention maps from our model. We also compare these maps with attention maps created by our Soft-Atten models from RGB and flow streams in Fig~\ref{fig:vis}. Notice that these two attention maps are qualitatively different across all methods. The appearance attention is likely to cover foreground objects or actors, while the motion attention focuses on the moving parts. Moreover, the appearance attention from our model can better localize the foreground regions of actions than those of Soft-Atten from the RGB stream, while the motion attention from our model remains similar to the Soft-Atten from the flow stream. We also find that the attention maps from our model are more ``diffused''. This is because the regularization by a uniform distribution in Prob-Atten leads to smoother attention maps.

\section{Conclusions}
In this paper, we presented a novel method of attention distillation for action recognition. We provided extensive experiments to evaluate our method. Our results demonstrate that proper design of the attention module helps to improve recognition performance. In addition, attention maps from RGB and flow networks are qualitatively different, suggesting that these networks capture different aspects of the video. We also showed that our method achieves competitive results for action recognition across datasets, and that attention distillation is more robust for learning a motion-aware video representation. We believe our work provides valuable insights into attention based recognition, and a step towards learning spatio-temporal video features via knowledge distillation.

\noindent \textbf{Acknowledgments}. Portions of this research were supported in part by National Science Foundation Award 1936970 and a gift from Facebook. YL acknowledges the support from the Wisconsin Alumni Research Foundation. XC acknowledges the support from Midea Emerging Technology Co., Ltd.
\bibliography{egbib}

\newpage
This is the supplementary material for our paper in BMVC 2020, titled `Attention Distillation for Learning Video Representations''. In this document, we introduce the implementation details. Moreover, we investigate the predicted attention maps and the learned features of our model and provide further analysis of our approach.

\section*{Network Architecture}
We detail the network architecture of our full model (Prob-Atten) in Table~\ref{table:structure}. Specifically, our model adopts I3D network~\cite{carreira2017quo} as the backbone. We attached the attention modules to the last Inception Module of the fourth convolution block. The appearance and motion attention maps, predicted by attention modules, are further used to pool features from the last Inception Module of the fifth convolution block for classification. And their results are fused at the end for final recognition. The network takes 24 frames as inputs with dimension $24\times224\times224\times3$ (RGB). And the network outputs (a) a downsampled attention map with size $3\times7\times7$ (three temporal slices with spatial resolution of $7\times 7$); and (b) the action scores for each category.

\section*{Analysis of Attention Distillation}
We provide extensive analysis to understand what has been learned by our model. We show that these attention maps help to locate the spatial extents of actions. And, we study different approaches to evaluate whether the learned representation is sensitive to motion. Finally, we provide more visualization of the attention maps of our model.\\

\noindent \textbf{Does the attention help to localize actions?} We evaluate our output attention for action localization using THUMOS'13 localization dataset~\cite{idrees2017thumos}--a subset of UCF101 with bounding box annotations for actions. We present our evaluation metric and discuss our results.
\begin{itemize}[leftmargin=*]
    \item \textbf{Evaluation Metric}.\ We consider action localization as binary labeling of pixels and report the F1 score from Precision-Recall (PR) curve. Specifically, we first rescale both attention maps and video frames into a fixed resolution ($56\times56$). We then enumerate all thresholds and binarize the attention map. Each threshold defines a point on the PR curve. Given a binary attention map, a positive pixel is considered as a true positive if it is inside the bounding box, or it is within 10-pixel ``tolerance zone'' of the box. This tolerance is added to compensate for the reduced resolution of the attention map, as in~\cite{oquab2015object}. We report the best F1 score on the curve and its corresponding precision and recall.

    \item \textbf{Results}.\ We compare attention maps from our model to a set of baseline methods, including a fixed Gaussian distribution (center prior), a latest deep saliency model (DSS~\cite{hou2017deeply}), and our Soft-Atten (RGB/Flow). The results are shown in Table~\ref{table:Localization}. Our appearance attention beats the baselines of center prior and Soft-Atten (RGB), but is worse than Soft-Atten (flow). Our motion attention achieves the highest score among all methods that only receive action labels as supervision, and only under-performs DSS. We have to emphasis that directly comparing our results to DSS is unfair. DSS is trained with pixel-level annotations using external data and runs at the original video resolution, while our attention maps are trained using clip-level action labels and down-sampled both spatially (32x) and temporally (8x). These results suggest that our attention maps help to locate the spatial extent of actions. 
\end{itemize}

\begin{table}
\centering
\tablestyle{2pt}{1.5}
\setlength{\tabcolsep}{12pt} % Default value: 6pt
\renewcommand{\arraystretch}{1} % Default value: 1
\footnotesize
\begin{tabular}{c|ccc}
Method                                  & Prec        & Recall   & F1    \\ \hline 
%Uniform                                 & 15.7        & 100.0    & 27.2  \\
Gaussian (center prior)                 & 52.6        & 20.6     & 29.6  \\ 
Saliency Map (DSS~\cite{hou2017deeply}) & \textbf{51.2} & 47.7     & \textbf{49.4}  \\ 
Soft-Atten (RGB)                        & 33.8        & 40.5     & 36.9  \\ 
Soft-Atten (Flow)                       & 39.2        & 50.0     & 44.0  \\ 
Our Appearance                          & 31.5        & 52.1     & 39.2  \\ 
Our Motion                              & 36.3        & \textbf{62.6}     & 46.0  \\ 
\end{tabular}
\caption{Results of action localization using attention maps on THUMOS'13 localization test set~\cite{idrees2017thumos}. We report the best F1 score and its precision and recall. Our motion attention outperforms all baselines that are trained with only action labels.\vspace{-0.6em}}
\label{table:Localization}
\end{table}

\noindent \textbf{Does our method learn better motion representation?} We further study how the temporal order of the input video frames will affect the recognition performance. We conduct an experiment of classifying reverted videos as in~\cite{Xie_2018_ECCV,zhou2017temporal}. Specifically, we invert the frame order for all testing videos of UCF101 and HMDB51. We compare their recognition results with those from normal temporal order. If a model truly rely on motion representation for the recognition, this inversion will significantly decrease the recognition performance. We test the vanilla I3D RGB and flow models, as well as our model. And the results are presented in Table~\ref{table:arrow}. Not surprisingly, I3D flow model has the largest performance drop. In contrast, I3D RGB is barely affected by the reverted arrow of time. Our model has a performance drop that is larger than I3D RGB yet much smaller than I3D flow. This is consistent with our results on action recognition. Our model does not capture the same level of motion information as the flow network.\\ 

\begin{table}[t]
\centering
\tablestyle{2pt}{1.0}
\setlength{\tabcolsep}{5pt} % Default value: 6pt
\renewcommand{\arraystretch}{1} % Default value: 1
\footnotesize
\begin{tabular}{c|c|ccc}
\multirow{2}{*}{Dataset}        & \multirow{2}{*}{Method}   & \multicolumn{3}{c}{Mean Class Accuracy} \\
    &                           & Original   & Reverted &  Delta$\Delta$ \\ \hline 

\multirow{3}{*}{UCF101}         & I3D RGB  & 94.8 & 94.7     & 0.1  \\
                                & I3D flow & 94.0 & 89.9     & \textbf{4.1}  \\ 
                                & Ours     & 95.7 & 95.1     & 0.6   \\  \hline
\multirow{3}{*}{HMDB51}         & I3D RGB  & 70.9 & 70.2     & 0.7  \\
                                & I3D flow & 73.9 & 66.0     & \textbf{7.9} \\ 
                                & Ours     & 72.0 & 70.6     & 1.4  \\
\end{tabular}
\vspace{0.1em}
\caption{Inverting the arrow of time for action recognition. We train the models on normal samples, yet test them on videos with reversed temporal order. A large performance drop indicates that the model has to rely on motion information for the recognition.\vspace{-1.6em}}
\label{table:arrow}
\end{table}

\noindent \textbf{How is the motion encoded?}\ It is also possible that our model simply copies the motion attention map without encoding motion in the network. To eliminate this hypothesis, we experimented with training an RGB network that directly combines a reference motion attention map and its own appearance attention map for action recognition. The reference motion attention is produced by a flow network during both training and testing. And the rest of this network follows exactly the same architecture as our model. This model has an accuracy of $95.1\%/71.6\%$ on UCF101/HMDB51, under-performing our model by -$0.6\%$/-$0.4\%$ on UCF101/HMDB51. These results indicate that the distillation process not only generates motion attention maps, but also learns motion-aware representation.\\

% \noindent \textbf{What has been learned?}\ Our visualization and action localization experiment suggest that our model learns to locate moving regions from video frames. However, when we invert the temporal order of frames, our learned features are not as sensitive as those from flow network. These results illustrate a key challenge for learning motion-aware representations. How the model learns to {\it identify} moving regions is not necessarily the right representation to {\it encode} motion. This is the same pitfall faced by our work and many previous works~\cite{ng2016actionflownet,liu2018end}. And this challenge remains open. 

% \section{Conclusions}
% In this paper, we presented a novel method of attention distillation for action recognition in videos. We provided extensive experiments to evaluate our method. Our results demonstrate that a proper design of attention module helps to improve recognition performance. More importantly, attention maps from RGB and flow networks are qualitatively different, suggesting that these networks capture different aspects of the video. We also showed that our method achieves competitive results of action recognition across datasets and attention distillation is more robust for learning motion-aware video representation. We believe our work provides valuable insights into attention based recognition, and a solid step towards learning spatiotemporal features and general application of knowledge distillation in deep models. 

\begin{figure*}[t]
\centering
\includegraphics[width=0.99\linewidth]{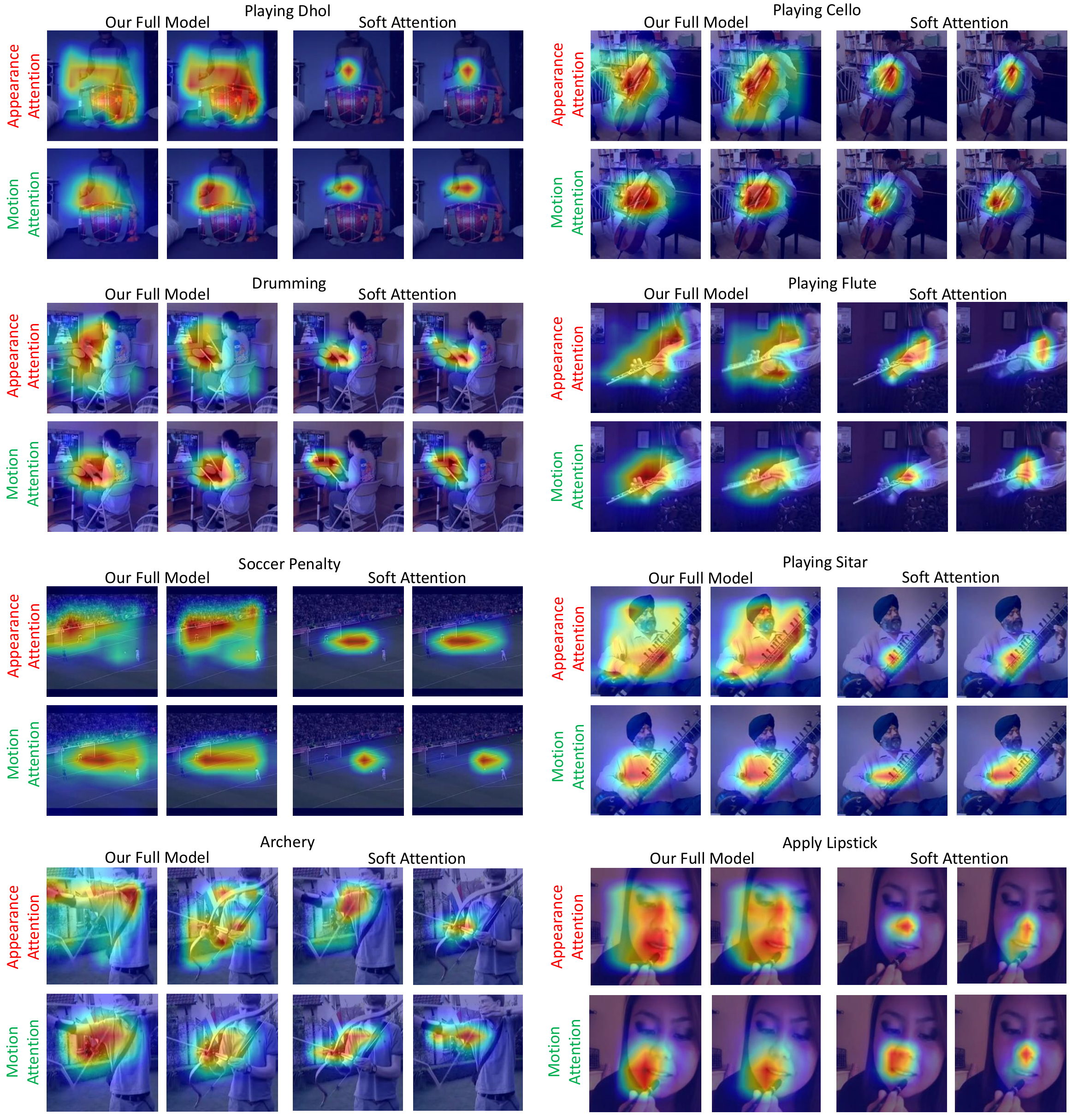}
\caption{Visualization of attention maps from our full model and Soft-Atten. For each 24 frames video clip, we plot the attention heatmap over the first frame and last frame. Our model produces qualitatively different appearance and motion attention maps. And these attention maps are better at localizing the actions when compared to vanilla soft attention. }
\label{fig:vis-more}
\vspace{-1em}
\end{figure*}
\noindent \textbf{Additional Visualizations}. We provide additional visualization of attention maps in Fig~\ref{fig:vis-more}. The figure follows the same format as Fig.\ 2 in our paper. These results further verify that (1) the appearance and motion attention maps are qualitatively different and (2) these attention map at good at localizing the actions, e.g., the actors or the moving regions.\\

\noindent \textbf{What has been learned?}\ Our visualization and action localization experiment suggest that our model learns to locate moving regions from video frames. However, when we invert the temporal order of frames, our learned features are not as sensitive as those from flow network. These results illustrate a key challenge for learning motion-aware representations. How the model learns to {\it identify} moving regions is not necessarily the right representation to {\it encode} motion. This is the same pitfall faced by our work and many previous work~\cite{ng2016actionflownet,liu2019end}. And this challenge remains open. 
\begin{table*}[t]
\fontsize{8}{10}\selectfont
\setlength{\tabcolsep}{1.1pt}
\centering
\label{table:network}
\begin{tabular}{c|c|c|c|c|c|c|c}
\hline 
\multirow{2}{*}{\textbf{ID}} & \multirow{2}{*}{\textbf{Branch}}                                                       & \multirow{2}{*}{\textbf{Type}}                                                & \multirow{2}{*}{\begin{tabular}[c]{@{}c@{}}\textbf{Kernel Size}\\ (THW)\end{tabular}} & \multirow{2}{*}{\begin{tabular}[c]{@{}c@{}}\textbf{Stride}\\ (THW)\end{tabular}} & \multirow{2}{*}{\begin{tabular}[c]{@{}c@{}}\textbf{Output Size}\\ (THWC)\end{tabular}} & \multirow{2}{*}{\textbf{Depth}} & \multirow{2}{*}{\textbf{Comments (Loss)}}                                                  \\
                   &                                                                               &                                                                      &                                                                              &                                                                         &                                                                               &                        &                                                                                  \\ \hline 
1                                       & \multirow{13}{*}{\begin{tabular}[c]{@{}c@{}}Backbone\\ (shared)\end{tabular}} & Convolution                                                          & 7x7x7                                                                        & 2x2x2                                                                   & 12x112x112x64                                                                 & 1                      &                                                                                  \\ \cline{3-8}
2                                       &                                                                               & Max Pool                                                             & 1x3x3                                                                        & 1x2x2                                                                   & 12x56x56x64                                                                   & 0                      &                                                                                  \\ \cline{3-8}
3                                       &                                                                               & Convolution                                                          & 1x1x1                                                                        & 1x1x1                                                                   & 12x56x56x64                                                                   & 1                      &                                                                                  \\ \cline{3-8}
4                                       &                                                                               & Convolution                                                          & 3x3x3                                                                        & 1x1x1                                                                   & 12x56x56x192                                                                  & 1                      &                                                                                  \\ \cline{3-8}
5                                       &                                                                               & Max Pool                                                             & 1x3x3                                                                        & 1x2x2                                                                   & 12x28x28x192                                                                  & 0                      &                                                                                  \\ \cline{3-8}
6                                       &                                                                               & Inception 3a                                                         &                                                                              &                                                                         & 12x28x28x256                                                                  & 2                      &                                                                                  \\ \cline{3-8}
7                                       &                                                                               & Inception 3b                                                         &                                                                              &                                                                         & 12x28x28x480                                                                  & 2                      &                                                                                  \\ \cline{3-8}
8                                       &                                                                               & Max Pool                                                             & 3x3x3                                                                        & 2x2x2                                                                   & 6x14x14x480                                                                   & 0                      &                                                                                  \\ \cline{3-8}
9                                       &                                                                               & Inception 4a                                                         &                                                                              &                                                                         & 6x14x14x512                                                                   & 2                      &                                                                                  \\ \cline{3-8}
10                                      &                                                                               & Inception 4b                                                         &                                                                              &                                                                         & 6x14x14x512                                                                   & 2                      &                                                                                  \\ \cline{3-8}
11                                      &                                                                               & Inception 4c                                                         &                                                                              &                                                                         & 6x14x14x512                                                                   & 2                      &                                                                                  \\ \cline{3-8}
12                                      &                                                                               & Inception 4d                                                         &                                                                              &                                                                         & 6x14x14x528                                                                   & 2                      &                                                                                  \\ \cline{3-8}
13                                      &                                                                               & Inception 4e                                                         &                                                                              &                                                                         & 6x14x14x832                                                                   & 2                      & Branching                                                                        \\ \hline 
14                                      & \multirow{6}{*}{\begin{tabular}[c]{@{}c@{}}Motion \\Attention \\ Branch\end{tabular}}       & \begin{tabular}[c]{@{}c@{}}Max Pool\\ (on Inception 4e)\end{tabular} & 2x3x3                                                                        & 2x2x2                                                                   & 3x7x7x832                                                                     & 0                      &                                                                                  \\ \cline{3-8}
15                                      &                                                                               & Convolution                                                          & 1x3x3                                                                        & 1x1x1                                                                   & 3x7x7x128                                                                     & 1                      &                                                                                  \\ \cline{3-8}
16                                      &                                                                               & Convolution                                                          & 1x1x1                                                                        & 1x1x1                                                                   & 3x7x7x1                                                                       & 1                      & \begin{tabular}[c]{@{}c@{}}KL Loss\\ (Attention Distillation)\end{tabular}              \\ \cline{3-8} 
17                                      &                                                                               & \begin{tabular}[c]{@{}c@{}}Gumbel Softmax\\ (Sampling)\end{tabular}                                                        &                                                                              &                                                                         & 3x7x7x1                                                                       & 0                      &      Sampling Attention Map                                                                            \\ \hline
18                                      & \multirow{6}{*}{\begin{tabular}[c]{@{}c@{}}Appearance \\Attention \\ Branch\end{tabular}}       & \begin{tabular}[c]{@{}c@{}}Max Pool\\ (on Inception 4e)\end{tabular} & 2x3x3                                                                        & 2x2x2                                                                   & 3x7x7x832                                                                     & 0                      &                                                                                  \\ \cline{3-8}
19                                      &                                                                               & Convolution                                                          & 1x3x3                                                                        & 1x1x1                                                                   & 3x7x7x128                                                                     & 1                      &                                                                                  \\ \cline{3-8}
20                                      &                                                                               & Convolution                                                          & 1x1x1                                                                        & 1x1x1                                                                   & 3x7x7x1                                                                       & 1                      & \begin{tabular}[c]{@{}c@{}}KL Loss\\ (Regularization)\end{tabular}              \\ \cline{3-8} 
21                                      &                                                                               & \begin{tabular}[c]{@{}c@{}}Gumbel Softmax\\ (Sampling)\end{tabular}                                                        &                                                                              &                                                                         & 3x7x7x1                                                                       & 0                      &      Sampling Attention Map                                                                            \\ \hline
22                                      & \multirow{8}{*}{\begin{tabular}[c]{@{}c@{}}Motion \\Action \\ Branch \end{tabular}}      & \begin{tabular}[c]{@{}c@{}}Max Pool\\ (on Inception 4e)\end{tabular} & 2x2x2                                                                        & 2x2x2                                                                   & 3x7x7x832                                                                     & 0                      &                                                                                  \\ \cline{3-8}
23                                      &                                                                               & Inception 5a                                                         &                                                                              &                                                                         & 3x7x7x832                                                                     & 2                      &                                                                                  \\ \cline{3-8} 
24                                      &                                                                               & Inception 5b                                                         &                                                                              &                                                                         & 3x7x7x1024                                                                    & 2                      &                                                                                  \\ \cline{3-8} 
25                                      &                                                                               & \begin{tabular}[c]{@{}c@{}}Weighted\\ Avg Pool\end{tabular}          & 2x7x7                                                                        & 1x1x1                                                                   & 2x1x1x1024                                                                    & 0                      & \begin{tabular}[c]{@{}c@{}}Weights from Gumbel Softmax\\ (Motion Attention Map)\end{tabular}     \\ \cline{3-8} 
26                                      &                                                                               & Fully Connected                                                      &                                                                              &                                                                         & 2x1x1x101                                                                     & 1                      &                                                                                  \\ \cline{3-8} 
27                                     &                                                                               & Avg Pool                                                             & 2x1x1                                                                        & 1x1x1                                                                   & 1x1x1x101                                                                     & 0                      &                                                                                  \\ \cline{3-8} 
28                                      &                                                                               & Softmax                                                              &                                                                              &                                                                         & 1x1x1x101                                                                     & 0                      & \begin{tabular}[c]{@{}c@{}}Cross Entropy Loss\\ (Action Recognition)\end{tabular} \\ \hline
29                                      & \multirow{8}{*}{\begin{tabular}[c]{@{}c@{}}Appearance \\ Action\\ Branch\end{tabular}}      & \begin{tabular}[c]{@{}c@{}}Max Pool\\ (on Inception 4e)\end{tabular} & 2x2x2                                                                        & 2x2x2                                                                   & 3x7x7x832                                                                     & 0                      &                                                                                  \\ \cline{3-8}
30                                      &                                                                               & Inception 5a                                                         &                                                                              &                                                                         & 3x7x7x832                                                                     & 2                      &                                                                                  \\ \cline{3-8} 
31                                      &                                                                               & Inception 5b                                                         &                                                                              &                                                                         & 3x7x7x1024                                                                    & 2                      &                                                                                  \\ \cline{3-8} 
32                                      &                                                                               & \begin{tabular}[c]{@{}c@{}}Weighted\\ Avg Pool\end{tabular}          & 2x7x7                                                                        & 1x1x1                                                                   & 2x1x1x1024                                                                    & 0                      & \begin{tabular}[c]{@{}c@{}}Weights from Gumbel Softmax\\ (Appearance Attention Map)\end{tabular}     \\ \cline{3-8} 
33                                      &                                                                               & Fully Connected                                                      &                                                                              &                                                                         & 2x1x1x101                                                                     & 1                      &                                                                                  \\ \cline{3-8} 
34                                     &                                                                               & Avg Pool                                                             & 2x1x1                                                                        & 1x1x1                                                                   & 1x1x1x101                                                                     & 0                      &                                                                                  \\ \cline{3-8} 
35                                      &                                                                               & Softmax                                                              &                                                                              &                                                                         & 1x1x1x101                                                                     & 0                      & \begin{tabular}[c]{@{}c@{}}Cross Entropy Loss\\ (Action Recognition)\end{tabular} \\ \hline

\end{tabular}
\caption{Network Architecture of our full model (Prob-Atten). The network attaches appearance and motion attention moduels to a backbone I3D network. We list details of all operations in the network, as well as where the loss functions are attached. Note that the predicted scores from Motion Action Branch and Appearance Action Branch are fused at the end for final recognition.}
\label{table:structure}
\end{table*}

\end{document}